\documentclass{article}

\usepackage{microtype}
\usepackage{graphicx}
\usepackage{subfigure}
\usepackage{booktabs} 
\usepackage{framed}
\usepackage{nicefrac}

\usepackage{hyperref}

\usepackage[accepted]{icml2024}

\usepackage{amsmath}
\usepackage{amssymb}
\usepackage{mathtools}
\usepackage{amsthm}

\usepackage[capitalize,noabbrev]{cleveref}

\theoremstyle{plain}
\newtheorem{theorem}{Theorem}[section]

\theoremstyle{definition}

\theoremstyle{remark}

\usepackage[textsize=tiny]{todonotes}
\usepackage{balance}

\icmltitlerunning{Tight Bounds for Online Convex Optimization with Adversarial Constraints}
\begin{document}

\twocolumn[
\icmltitle{Tight Bounds for Online Convex Optimization with Adversarial Constraints}




\begin{icmlauthorlist}
\icmlauthor{Abhishek Sinha}{yyy}
\icmlauthor{Rahul Vaze}{yyy}
\end{icmlauthorlist}

\icmlaffiliation{yyy}{School of Technology and Computer Science, Tata Institute of Fundamental Research, Mumbai 400005, India}

\icmlcorrespondingauthor{Abhishek Sinha}{abhishek.sinha@tifr.res.in}

\icmlkeywords{Online Convex Optimization, Regret bounds}

\vskip 0.3in
]



\printAffiliationsAndNotice{} 

\begin{abstract}
 A well-studied generalization of the standard online convex optimization (OCO) is constrained online convex optimization (COCO). In COCO, on every round, a convex cost function and a convex constraint function are revealed to the learner after the action for that round is chosen. The objective is to design an online policy that simultaneously achieves a small regret while ensuring small cumulative constraint violation (CCV) against an adaptive adversary. A long-standing open question in COCO is whether an online policy can simultaneously achieve $O(\sqrt{T})$ regret and $O(\sqrt{T})$ CCV without any restrictive assumptions. For the first time, we answer this in the affirmative and show that an online policy can simultaneously achieve $O(\sqrt{T})$ regret and $\tilde{O}(\sqrt{T})$ CCV. We establish this result by effectively combining the adaptive regret bound of the AdaGrad algorithm with Lyapunov optimization - a classic tool from control theory. Surprisingly, the analysis is short and elegant.
\end{abstract}
\section{Introduction} \label{intro}
We consider the Constrained Online Convex Optimization problem (COCO for short) with adversarially chosen convex cost and convex constraint functions. 
In COCO, on every round, a convex cost function and a convex constraint function are revealed to the learner, after the action for that round is chosen. The objective is to design an online policy that simultaneously achieves a small regret (w.r.t. cost functions) while ensuring a small cumulative constraint violations (CCV) (w.r.t. constraint functions) against any adaptive adversary that is allowed to choose a single action across all rounds. 

The COCO problem has been extensively studied in the literature over the past decade and the best-known bounds for simultaneouls regret and CCV  are $O(\sqrt{T})$ and $O(T^{\nicefrac{3}{4}})$, respectively \citep{guo2022online, yi2023distributed, sinha2023playing}, without any additional assumptions.  A simultaneous lower bound on regret and CCV is known to be 
$\Omega(\sqrt{T})$ and $\Omega(\sqrt{T})$ \citep{sinha2023playing}. Please refer to Table \ref{comp-table} in the Appendix for a summary of the major results.

It is known that with additional assumptions, simultaneous $O(\sqrt{T})$ regret and $O(\sqrt{T})$ CCV is achievable (See the Related Work Section below). However, even after more than a decade of sustained work on this problem, the question: whether an online policy can simultaneously achieve $O(\sqrt{T})$ regret and $O(\sqrt{T})$ CCV, without any additional assumptions remains unresolved  and constitutes a
major open problem in this area. 

In this paper, we resolve this problem for the first time 
and make the following contributions. 
\begin{enumerate}
	\item 
We propose an efficient policy that simultaneously achieves $O(\sqrt{T})$ regret and $O(\sqrt{T}\log T)$ CCV for the COCO problem. Our policy breaks the long-standing $O(T^{\nicefrac{3}{4}})$ barrier for the CCV and achieves the lower bound up to a logarithmic term. 
\item The proposed policy simply runs the standard AdaGrad algorithm on a specially constructed sequence of convex surrogate cost functions and requires no explicit parameter tuning. On every round, the policy computes only a gradient and performs an Euclidean projection. 
\item Unlike the primal-dual algorithms in the previous works, the design of our policy employs the Lyapunov method from the classical control theory. The presented analysis is short and elegant, may be of independent interest for other related problems.
\end{enumerate}

\section{Related Work}
While the regret bound of $O(\sqrt{T})$ for COCO is minimax optimal, a number of papers have obtained tighter CCV bounds  under various assumptions, such as time-invariant constraints \citep{mahdavi2012trading, jenatton2016adaptive, yuan2018online, yu2020low, yi2021regret}, stochastic constraints \citep{yu2017online}, Slater's condition \citep{neely2017online, yi2023distributed}, or the non-negativity of the regret \citep{sinha2023playing}.
Furthermore, the algorithms proposed in \citet{guo2022online, yi2023distributed, yi2021regret, yu2020low} are complex as they need to solve a full-fledged convex optimization problem on each round. 

\section{Problem formulation} 
Consider a repeated game between an online policy and an adaptive adversary. In this game, on each round $t\geq 1,$ the policy chooses a feasible action $x_t$ from an admissible set $\mathcal{X}$. The set $\mathcal{X}$ is assumed to be non-empty, closed, and convex with a finite Euclidean diameter $D.$ Upon observing the current action $x_t,$ the adversary chooses two convex functions - a \emph{cost} function $f_t: \mathcal{X} \to \mathbb{R}$ and a \emph{constraint} function $g_t: \mathcal{X} \to \mathbb{R}.$ The constraint function $g_t$ correspond to an online constraint of the form $g_t(x)\leq 0.$ Since the constraints are adaptively chosen, they cannot be expected to be satisfied by the policy on each round. Consequently, the policy incurs a cost of $f_t(x_t)$ and an instantaneous constraint violation of $\max(0, g_t(x_t)).$ Our high-level objective is to design an online policy that achieves a small cumulative cost while approximately satisfying the constraints in the long term. Throughout the paper, we assume that all cost and constraint functions are convex and $G$-Lipschitz.
The regret of any policy is computed by comparing its cumulative cost against that of a fixed feasible action $x^\star \in \mathcal{X}$ that satisfies all constraints on each rounds. Specifically, let $\mathcal{X}^\star \subseteq \mathcal{X}$ be the set consisting of all actions satisfying all constraints:
\begin{eqnarray} \label{feas-set}
	\mathcal{X}^\star = \{x \in \mathcal{X}: g_t(x) \leq 0, ~\forall t \geq 1\}.
\end{eqnarray} 
We assume that the feasible set is non-empty, \emph{i.e.,} $\mathcal{X}^\star \neq \emptyset.$
We then define the Regret and the CCV of any policy as 
\begin{eqnarray}
	\textrm{Regret}_T &=& \sup_{x^\star \in \mathcal{X}^\star} \sum_{t=1}^T\big(f_t(x_t)-f_t(x^\star)\big), \label{reg-def}\\
	\textrm{CCV}_T&=& \sum_{t=1}^T (g_t(x_t))^+, \label{ccv-def}
\end{eqnarray}
where we have defined $(y)^+\equiv \max(0,y), ~ y \in \mathbb{R}.$ 

In the standard OCO problem, only the cost functions are revealed on every round and there is no online constraint function. The goal for the standard OCO problem is to only minimize the regret.
Hence, the standard OCO problem can be seen to be a special case of COCO where $g_t=0, \forall t,$ and hence, $\mathcal{X}^\star=\mathcal{X}$.


\paragraph{Note:} 
We may consider an extension of the above problem where $k>1$ constraints of the form $g_{t,i}(x) \leq 0, i\in [k],$ are revealed on each round. These multiple constraints can be combined to a single constraint 
by replacing them with a new constraint $g_t(x)\leq 0,$ where the function $g_t$ is taken to be the point wise max, \emph{i.e.,} 
 $g_t(x) \equiv \max_{i=1}^k g_{t,i}(x).$


\section{Optimal Regret and Constraint Violation Bounds for COCO} \label{tighter_bds}

\subsection{Overview of the technique}
Compared to the standard online convex optimization problem where the objective is to minimize the Regret \citep{hazan2022introduction}, in this problem our objective is two fold - to \emph{simultaneously} control the Regret and the CCV. Towards this end, in the following, we 
propose a Lyapunov-based technique that 
yields $O(\sqrt{T})$ regret and $\tilde{O}(\sqrt{T})$ CCV. 
\subsection{Preliminaries}
We now briefly recall the first-order methods (\emph{a.k.a.} Projected Online Gradient Descent (OGD)) for the standard OCO problem. These methods differ among each other in the way the step sizes are chosen. For a sequence of convex cost functions $\{\hat{f}_t\}_{t \geq 1},$ a projected OGD algorithm selects the successive actions as \citep[Algorithm 2.1]{orabona2019modern}:
\begin{eqnarray}\label{ogd-policy} 
	x_{t+1} = \mathcal{P}_\mathcal{X}(x_t - \eta_t \nabla_t), ~~ \forall t\geq 1,
\end{eqnarray}
where $\nabla_t \equiv \nabla \hat{f}_t(x_t)$ is a subgradient of the function $\hat{f}_t$ at $x_t$, $\mathcal{P}_\mathcal{X}(\cdot)$ is the Euclidean projection operator on the set $\mathcal{X}$ and $\{\eta_t\}_{t \geq 1}$ is a specified step size sequence. 
The (diagonal version of the) AdaGrad policy adaptively chooses the step size sequence as a function of the previous subgradients as  $\eta_t= \frac{\sqrt{2}D}{2\sqrt{\sum_{\tau=1}^{t} G_\tau^2}},$ where $G_t=||\nabla_t||_2, t \geq 1$ \citep{duchi2011adaptive}. \footnote{We set $\eta_t=0$ if $G_t=0.$} It enjoys the following adaptive regret bound.
\begin{theorem}{\citep[Theorem 4.14]{orabona2019modern}}  The AdaGrad policy, with the above step size sequence, achieves the following regret bound for the standard OCO problem: 
	\begin{eqnarray} \label{cvx-reg-bd}
			 \textrm{Regret}_T \leq \sqrt{2}D \sqrt{\sum_{t=1}^T G_t^2}.
	\end{eqnarray}
	\end{theorem}

\subsection{Design and Analysis of the Algorithm}
 
To simplify the analysis, we pre-process the cost and constraint functions as follows.

\vspace{5pt}

\hrule
\textbf{Pre-processing:}
On every round, we clip the negative part of the constraint to zero by passing it through the standard ReLU unit. Next, we scale both the cost and constraint functions by a factor of $\alpha \equiv (2GD)^{-1}.$  More precisely, 
 we define $\tilde{f}_t \gets \alpha f_t, \tilde{g}_t \gets \alpha (g_t)^+.$ Hence, the pre-processed functions are $\alpha G=(2D)^{-1}$-Lipschitz with $\tilde{g}_t \geq 0, \forall t.$  
 \vspace{5pt}

\hrule 
In the following, we will derive the Regret and CCV bounds for the pre-processed functions. The corresponding bounds for the original unscaled functions are obtained upon scaling the results back by $\alpha^{-1}$ in the final step.
\subsubsection{Surrogate cost functions} 

Let $Q(t)$ denote the CCV for the pre-processed constraints up to round $t.$ Clearly, $Q(t)$ satisfies the simple recursion $Q(t)=Q(t-1)+\tilde{g}_t(x_t), t\geq 1, $ with $Q(0)=0$. Recall that one of our objectives is to make $Q(t)$ as small as possible.
Towards this, let $\Phi: \mathbb{R}_+ \to \mathbb{R}_+$ be any non-decreasing convex potential (Lyapunov) function such that $\Phi(0)=0.$ Using the convexity of the function $\Phi(\cdot),$ we have
%
\begin{eqnarray} \label{dr-bd-gen}
	\Phi(Q(t))  &\leq& \Phi(Q(t-1)) + \Phi'(Q(t))(Q(t)-Q(t-1)) \nonumber \\
&=& \Phi(Q(t-1)) + \Phi'(Q(t)) \tilde{g}_t(x_t). 
\end{eqnarray}
Hence, the one-step change (\emph{drift}) of the potential function $\Phi(Q(t))$ can be upper bounded as 
\begin{eqnarray} \label{drift_ineq_new}
	\Phi(Q(t))-\Phi(Q(t-1)) \leq \Phi'(Q(t)) \tilde{g}_t(x_t). 
\end{eqnarray}
Recall that, in addition to controlling the CCV, we also want to minimize the cumulative cost (which is equivalent to the regret minimization). Inspired by the stochastic \emph{drift-plus-penalty} framework of \citet{neely2010stochastic}, we combine these two objectives to a single objective of minimizing a sequence of surrogate cost functions $\{\hat{f}_t\}_{t=1}^T$ which is obtained by taking a positive linear combination of the drift upper bound \eqref{drift_ineq_new} and the cost function. More precisely, we define
\begin{eqnarray} \label{surrogate_new}
	\hat{f}_t(x):= V\tilde{f}_t(x)+ \Phi'(Q(t)) \tilde{g}_t(x), ~~ t \geq 1. 
\end{eqnarray}
In the above, $V$ is a suitably chosen non-negative parameter to be fixed later. Our proposed policy for COCO, described in Algorithm \ref{coco_alg}, simply runs the AdaGrad algorithm on the surrogate cost function sequence  $\{\hat{f}_t\}_{t\geq 1}$, for a specific choice of $\Phi(\cdot)$ and $V$ as dictated by the following analysis.
\begin{algorithm}[tb]
   \caption{Online Policy for COCO}
   \label{coco_alg}
\begin{algorithmic}[1]
   \STATE {\bfseries Input:} Horizon length $T,$ Sequence of convex cost functions $\{f_t\}_{t=1}^T$ and constraint functions $\{g_t\}_{t=1}^T,$ an upper bound $G$ to the norm of their (sub)-gradients, Diameter $D$ of the admissible set $\mathcal{X}$
     \STATE {\bfseries Parameters:} $V=1, \lambda = 1/(2\sqrt{T}), \Phi(x)= \exp(\lambda x)-1, \alpha=(2GD)^{-1}.$
  \STATE {\bfseries Initialization:} Set $ x_1=0, Q(0)=0$.
   \FOR{$t=1:T$}
   \STATE Choose $x_t,$ observe $f_t, g_t,$ incur a cost of $f_t(x_t)$ and constraint violation of $(g_t(x_t))^+$
   \STATE $\tilde{f}_t \gets \alpha f_t, \tilde{g}_t \gets \alpha \max(0,g_t).$
   \STATE $Q(t)=Q(t-1)+\tilde{g}_t(x_t).$
   \STATE Compute $\nabla_t = \nabla \hat{f}_t(x_t),$ where $\hat{f}_t$ is defined in \eqref{surrogate_new}
   \STATE $x_{t+1} = \mathcal{P}_\mathcal{X}(x_t - \eta_t \nabla_t)$, where $\eta_t= \frac{\sqrt{2}D}{2\sqrt{\sum_{\tau=1}^{t} ||\nabla_\tau||_2^2}}.$
   \ENDFOR
\end{algorithmic}
\end{algorithm}

\subsubsection{The Regret Decomposition Inequality}
Let $x^\star \in \mathcal{X}^\star$ be any feasible action as defined in \eqref{feas-set}. Using the drift inequality from Eqn. \eqref{drift_ineq_new}, the definition of surrogate costs \eqref{surrogate_new}, and the fact that $g_\tau(x^\star)\leq 0, \forall \tau \geq 1,$ we have
\begin{eqnarray*}
	&&\Phi(Q(\tau))-\Phi(Q(\tau-1)) + V(\tilde{f}_\tau(x_\tau)-\tilde{f}_\tau(x^\star)) \\
	&\leq& \hat{f}_\tau(x_\tau) - \hat{f}_\tau(x^\star), ~ \forall \tau \geq 1.
\end{eqnarray*}
Summing up the above inequalities for $1\leq \tau \leq t$, and using the fact that $\Phi(0)=0,$ we obtain  
\begin{eqnarray} \label{gen-reg-decomp}
	\Phi(Q(t)) +V \textrm{Regret}_t(x^\star) \leq \textrm{Regret}_t'(x^\star), ~ \forall x^\star \in \Omega^\star,
\end{eqnarray}
where $\textrm{Regret}_t$ on the LHS and $\textrm{Regret}'_t$ on the RHS of \eqref{gen-reg-decomp} refers to the regret for learning the pre-processed cost functions $\{\tilde{f}_t\}_{t\geq 1}$ and the surrogate cost functions $\{\hat{f}_t\}_{t \geq 1}$ respectively.
Since Algorithm \ref{coco_alg} uses the AdaGrad algorithm for learning the surrogate cost functions, from \eqref{cvx-reg-bd}, we need to upper bound the gradients of the surrogate functions to derive the regret bound. Towards this, the $\ell_2$-norm of the gradients $G_t$ of the surrogate cost function $\hat{f}_t$ can be bounded using the triangle inequality as follows:
\begin{eqnarray} \label{grad_bd_new}
	G_t
	&\leq& V||\nabla \tilde{f}_t(x_t)||_2+ \Phi'(Q(t))||\nabla \tilde{g}_t(x_t)||_2 \nonumber\\
	&\leq& (2D)^{-1}\big(V+\Phi'(Q(t)\big).
\end{eqnarray}
where in the last step, we have used the fact that the pre-processed functions are $(2D)^{-1}$-Lipschitz. 
Hence, plugging in the adaptive regret bound \eqref{cvx-reg-bd} on the RHS of \eqref{gen-reg-decomp}, we arrive at the following generalized regret decomposition inequality valid for any $t \geq 1: $
\begin{eqnarray} \label{gen-fn-ineq}
		\Phi(Q(t)) +V \textrm{Regret}_t(x^\star) \leq \sqrt{\sum_{\tau=1}^t \big(\Phi'(Q(\tau))\big)^2} + V\sqrt{t}.
\end{eqnarray}
where we have utilized simple algebraic inequalities $(x+y)^2 \leq 2(x^2+y^2)$ and $\sqrt{a+b} \leq \sqrt{a} + \sqrt{b}, a, b\geq 0.$ 

Finally, recall that the sequence $\{Q(t)\}_{t\geq 1}$ is non-negative and non-decreasing as $\tilde{g}_t\geq 0.$ Furthermore, the derivative $\Phi'(\cdot)$ is non-decreasing as the function $\Phi$ is assumed to be convex. Hence, upper-bounding all terms in the summation of the RHS of \eqref{gen-fn-ineq} by the last term, we arrive at the following simplified bound
\begin{eqnarray} \label{gen-fn-ineq2}
	\Phi(Q(t)) +V \textrm{Regret}_t(x^\star) \leq \Phi'\big(Q(t)\big)\sqrt{t} + V\sqrt{t}.
\end{eqnarray}
The regret decomposition inequality \eqref{gen-fn-ineq2} constitutes the key step for the subsequent analysis.
\subsubsection{Analysis}
\paragraph{An Exponential Lyapunov function:} We now derive the Regret and CCV bounds for the proposed  policy with an exponential Lyapunov function $\Phi(x)\equiv \exp(\lambda x)-1,$ for some parameter $\lambda \geq 0$ to be fixed later. An analysis with a power-law potential function is given in Appendix \ref{power-law}, which also yields similar bounds. It is easy to verify that the function $\Phi(\cdot)$ satisfies required conditions- it is a non-decreasing and convex function with $\Phi(0)=0.$ 
\paragraph{Bounding the Regret:}
From \eqref{gen-fn-ineq2}, for any feasible $x^\star \in \mathcal{X}^\star$ and for any $t \in [T],$ we have
\begin{eqnarray*}
	\exp(\lambda Q(t)) + V \textrm{Regret}_t(x^*) \leq \lambda \exp(\lambda Q(t)) \sqrt{t} + V\sqrt{t}.
\end{eqnarray*}
Transposing, this yields the following bound on Regret: 
\begin{eqnarray} \label{reg-bd-exp}
	\textrm{Regret}_t(x^\star ) \leq \sqrt{t}+\frac{1}{V}+ \frac{\exp(\lambda Q(t))}{V}(\lambda \sqrt{t}-1).
\end{eqnarray}
Choosing any $\lambda \leq  \frac{1}{\sqrt{T}},$ the last term in the above inequality becomes non-positive for any $t \in [T].$ Hence, for any $x^\star \in \mathcal{X}^\star$, we have 
\begin{eqnarray} \label{regret-bd1}
	\textrm{Regret}_t(x^\star ) \leq \sqrt{t}+\frac{1}{V}. ~~ \forall t \in [T].
\end{eqnarray}
\paragraph{Bounding the CCV:} 
Since all pre-processed cost functions are $(2D)^{-1}$-Lipschitz,
we trivially have $\textrm{Regret}_t(x^\star)\geq -\frac{Dt}{2D} \geq -\frac{t}{2}.$ Hence, from Eqn.\ \eqref{reg-bd-exp}, we have for any $\lambda < \frac{1}{\sqrt{T}}$ and any $t \in [T]:$
\begin{eqnarray} \label{q-len-exp-bd}
	&&\frac{\exp(\lambda Q(t))}{V}(1-\lambda \sqrt{t}) \leq  2t + \frac{1}{V} \nonumber \\
	 &&\implies  Q(t) \leq  \frac{1}{\lambda}\ln\frac{1+2Vt}{1-\lambda \sqrt{t}}.
\end{eqnarray}
Finally, setting $\lambda=\frac{1}{2\sqrt{T}}$ and $V=1,$ and scaling the bounds back by $\alpha^{-1}\equiv 2GD,$ we arrive at our main result.
\begin{theorem} \label{main_result}
For the COCO problem with adversarially chosen $G$-Lipschitz cost and constraint functions, Algorithm \ref{coco_alg} yields the following Regret and CCV bounds for any horizon length $T \geq 1:$
\begin{eqnarray*}
 \textrm{Regret}_t &\leq& 2GD(\sqrt{t}+1), ~~\forall t \in [T]\\ \textrm{CCV}_T &\leq& 4GD\ln(2\big(1+2T)\big)\sqrt{T}.
 \end{eqnarray*}
 In the above, $D$ denotes the Euclidean diameter of the closed and convex admissible set $\mathcal{X}$.
\end{theorem}
\paragraph{Remarks:} Although, in the above, we assume that the horizon length $T$ is known, we can use the standard doubling trick for unknown $T.$

In the final section, we show that in addition to a small regret and CCV bounds, the proposed Algorithm \ref{coco_alg} also has a small movement cost measured in the (squared) Euclidean metric. These features make it attractive for practical applications.

\subsection{Bounding the movement cost}

The movement cost of the AdaGrad policy can be bounded as
\begin{eqnarray*}
	||x_{t+1}-x_{t}||_2^2 &\stackrel{(a)}{\leq}& ||\eta_t \nabla_t||_2^2 
	= \frac{D^2}{2} \frac{G_t^2}{\sum_{\tau=1}^t G_\tau^2}\\ &\leq& \frac{D^2}{2} \int_{\sum_{\tau=1}^{t-1}G_\tau^2}^{\sum_{\tau=1}^{t}G_\tau^2} \frac{dx}{x},
\end{eqnarray*}
where in step (a), we have used the Pythagorean theorem for Euclidean projection \citep[Theorem 4]{hazan2022introduction}. 
Summing up, the cumulative squared-Euclidean movement cost of the AdaGrad policy for $T$ rounds can be bounded as 
\begin{eqnarray} \label{sq_movement_cost}
	\sum_{t=1}^{T} ||x_{t+1}-x_t||_2^2 
	&\leq& \frac{D^2}{2} \big(1+\ln (\sum_{\tau=1}^T G_\tau^2) - \ln G_1^2 \big).~~~ \hspace{10pt}
\end{eqnarray}
Equation \eqref{sq_movement_cost} gives an upper bound to the movement cost for the AdaGrad policy for any sequence of cost functions. 

\subsubsection{Movement cost of Algorithm \ref{coco_alg}}
Since Algorithm \ref{coco_alg} uses the AdaGrad policy on the sequence of surrogate cost functions, 
we can use Eqn.\ \eqref{sq_movement_cost} for bounding its total movement cost. Recall that, Eqn.\ \eqref{grad_bd_new} gives an upper bound to the gradient of the surrogate cost functions fed to the AdaGrad policy. Hence, we have
\begin{eqnarray}
\sum_{\tau=1}^T G_\tau^2 &\leq& G^2\sum_{\tau=1}^T (V+ \Phi'(Q(\tau))^2 \nonumber\\
&\stackrel{(a)}{\leq}& 2G^2T(1 + \lambda^2 \exp(2\lambda Q(T)) ) \nonumber\\
&\stackrel{(b)}{\leq} & 4G^2T(3+2T), \label{mv-cost}
\end{eqnarray}
where in (a), we have used the inequality $(x+y)^2 \leq 2(x^2+y^2),$ substituted the value of the parameter $V=1$ and used the fact that $\{Q(t\}_{t \geq 1}$ is a non-decreasing sequence. In (b), we have used the bound from \eqref{q-len-exp-bd} and substituted for the parameter $\lambda = 1/(2\sqrt{T}).$
Substituting the bound \eqref{mv-cost} into \eqref{sq_movement_cost}, the squared-Euclidean movement cost for Algorithm \ref{coco_alg} can be bounded as
		$\sum_{t=1}^{T} ||x_{t+1}-x_t||_2^2 = O((\log T)).$
Using the Cauchy-Schwarz inequality, the Euclidean movement cost can also be bounded as $\sum_{t=1}^{T} ||x_{t+1}-x_t||_2 \leq \sqrt{T} \sqrt{\sum_{t=1}^{T} ||x_{t+1}-x_t||^2_2}=\tilde{O}(\sqrt{T}).$

\section{Conclusion} \label{conclusion}
In this paper, we proposed an efficient online policy that achieves the optimal regret and CCV bounds for the COCO problem without any assumptions. We have also shown that the squared Euclidean movement cost of the proposed policy is small and is bounded by $O((\log T)^2).$ These results are established by using a new proof technique involving an exponential Lyapunov function. 

\clearpage
\balance
\bibliography{OCO.bib}
\bibliographystyle{icml2024}

\newpage
\appendix
\onecolumn
\section{Appendix}

\begin{table}[!ht]
\caption{Summary of the results for the COCO problem with convex cost and convex constraint functions. Unless specified otherwise in the Assumptions, both the cost and constraint functions are assumed to be chosen by an adaptive adversary. Details of the abbreviations appearing on the table can be found in the trailing note.}
\label{comp-table}
\vskip 0.15in
\begin{center}
\begin{small}
\begin{sc}
\begin{tabular}{lccccr}
\toprule
Reference & Regret & CCV & Complexity per round & Assumptions&\\
\midrule
\citet{mahdavi2012trading} & $O(\sqrt{T})$& $O(T^{\nicefrac{3}{4}})$& Proj.& Fixed Constr. \\\citet{jenatton2016adaptive}& $O(T^{\max(\beta, 1-\beta) })$& $O(T^{1- \beta/2})$& Proj& Fixed Constr.& \\
\citet{yu2017online}    & $O(\sqrt{T})$& $O(\sqrt{T})$& Proj& Slater \& Stoch. &       \\
\citet{neely2017online}     & $O(\sqrt{T})$ & $O(\sqrt{T})$& Proj.& Slater&\\
\citet{yu2020low} & $O(\sqrt{T})$ & $O(1)$& Conv-OPT & Slater \& Fixed Constr.&\\
\citet{yi2021regret}     & $O(\sqrt{T})$& $O(T^{\nicefrac{1}{4}})$& Conv-OPT& Fixed Constr.&\\
\citet{guo2022online}    & $O(\sqrt{T})$& $O(T^{\nicefrac{3}{4}})$& Conv-OPT& ---& \\
\citet{yi2023distributed}    & $O(T^{\max(\beta,1-\beta)})$& $O(T^{1-\beta})$& Conv-OPT& Slater &\\
\citet{sinha2023playing}    & $O(\sqrt{T})$& $O(T^{\nicefrac{3}{4}})$& Proj& ---&        \\
\citet{sinha2023playing}    & $O(\sqrt{T})$& $O(\sqrt{T})$& Proj& $\textrm{Regret}_T \geq 0$ &       \\
\textbf{This paper}   & $O(\sqrt{T})$& $\tilde{O}(\sqrt{T})$& Proj& --- &\\
\bottomrule
\end{tabular}
\end{sc}
\end{small}
\end{center}
\vskip -0.1in
\end{table}
\paragraph{Notes:}
\begin{enumerate}
	\item \textsc{Conv-OPT}: Solution of a convex optimization problem over the convex constraint set $\mathcal{X}$ 
	\item \textsc{Proj}: Euclidean projection onto the  set $\mathcal{X}$ 
	 \item $\textsc{Proj}$ operation is computationally much cheaper than \textsc{Conv-OPT} for typical $\mathcal{X}.$ 
	\item \textsc{Slater}: Slater condition
	\item \textsc{Stoch}: Stochastic constraints, \emph{i.e.,} the constraint functions are generated i.i.d. per slot.
	\item \textsc{Fixed Constr.}: Time-invariant fixed constraint functions known to the algorithm. Here the goal is to avoid the complex projection step onto the constraint set.
	\item $\beta$ is any fixed parameter in the interval $[0,1]$ 

\end{enumerate}

\subsection{Details on the assumptions made in the previous papers}
\paragraph{Slater's condition:}
This condition assumes that there exists a positive $\epsilon>0$ and a feasible action $x \in \mathcal{X}$ such that $g_t(x) \leq -\epsilon, \forall t \in [T].$ Trivially, Slater's condition does not hold for clipped constraints, \emph{i.e.}, constraints of the form $g_t(x)\equiv\max(0, \tilde{g}_t(x)),$ for some convex function $\tilde{g}_t.$ Worse yet, upon assuming Slater's condition, the CCV bounds derived in \citet[Theorem 4 (b)]{neely2017online}, \citet[Theorem 1]{yu2017online}, \citet{yu2020low}, and \citet[Theorem 2]{yi2023distributed} diverge to infinity as the slack $\epsilon$ approaches zero.  This should be compared against our main result in Theorem \ref{main_result}, where the bounds remain finite for any finite $G,D,$ and $T$ while making no assumption on Slater's condition. 

\paragraph{Non-negativity of Regret:} Since the benchmark action in the regret definition \eqref{reg-def} belongs to the feasible set $\mathcal{X}^\star,$ it is computationally non-trivial to determine whether the regret is non-negative or not \citep{sinha2023playing}. While it is shown in \citet[Theorem 9]{sinha2023playing}, that the improved $O(\sqrt{T})$ CCV bound holds for convex adversaries and fixed constraints, improving the CCV bound for the general case with adversarially chosen constraints and arbitrary adversaries was left as an open problem. 

\section{Analysis with A Power-law Lyapunov Function} \label{power-law}
We now specialize inequality \eqref{gen-fn-ineq2} by considering the power-law Lyapunov function $\Phi(x)\equiv x^n$ for some integer $n \geq 2,$ to be fixed later. The pseudocode for the policy remains the same as in Algorithm \ref{coco_alg} with the new set of parameters $ \alpha=\frac{1}{2GD}, n=\max(2, \lceil \ln T \rceil), V=(n-1)^{n-1}T^{\frac{n-1}{2}}, \Phi(x)=x^n.$ 

From \eqref{gen-fn-ineq2}, we obtain 
\begin{eqnarray} \label{fn-ineq2}
	Q^n(t)+V\textrm{Regret}_t(x^\star) \leq nQ^{n-1}(t) \sqrt{t} + V\sqrt{t}.
\end{eqnarray} 

\paragraph{Bounding the Regret:}
From Eqn.\ \eqref{fn-ineq2}, we have the following regret bound for any feasible $x^\star \in \mathcal{X}^\star:$
\begin{eqnarray*}
	\textrm{Regret}_t(x^\star) &\leq& \sqrt{t} + \frac{Q^{n-1}(t)}{V}(n\sqrt{t}-Q(t)) \\
	&\leq& \sqrt{t} + \frac{(n-1)^{n-1} t^{n/2}}{V},
\end{eqnarray*}
where the last step follows from the AM-GM inequality. Finally, upon taking $V=(n-1)^{n-1}T^{\frac{n-1}{2}}$ as in Algorithm \ref{coco_alg}, we obtain $\textrm{Regret}_t(x^\star) \leq 2\sqrt{t}, ~\forall 1\leq t \leq T$.
 
\paragraph{Bounding the CCV:} Since all pre-processed cost functions are $(2D)^{-1}$-Lipschitz,
we trivially have $\textrm{Regret}_t(x^\star)\geq -\nicefrac{Dt}{2D} \geq -\frac{t}{2}.$ Hence, from Eqn.\ \eqref{fn-ineq2}, we obtain
\begin{eqnarray} \label{qn-bd}
	Q^n(t) \leq 2Vt + nQ^{n-1}(t) \sqrt{t}.
\end{eqnarray}
Finally, to bound $Q(t)$ from the above, consider the case when $Q(t) \geq 2n\sqrt{t}$. In this case, from \eqref{qn-bd}, we have 
\begin{eqnarray*}
	Q^n(t) \leq 2Vt + \frac{1}{2}Q^n(t) \implies Q(t) \leq (4Vt)^{\frac{1}{n}}.
\end{eqnarray*}
Thus, in general, the following bound holds 
	\[ Q(t) \leq \max(2n\sqrt{t}, (4Vt)^{\frac{1}{n}}).\]
Substituting the chosen value for the parameter $V$ and using the fact that $n\geq 2$, from the above, we have 
\begin{eqnarray}\label{q-bd-new2}
Q(t)\leq \max (2n\sqrt{T}, 2nT^{\frac{1}{2}+ \frac{1}{2n}}) = 2nT^{\frac{1}{2}+ \frac{1}{2n}}, ~ \forall t \in [T].
\end{eqnarray} 
Finally, setting $n= \max(2,\lceil\ln T \rceil)$ as in Algorithm \ref{coco_alg}, we obtain the following CCV bound:
\begin{eqnarray*}
	\textrm{CCV}_T=Q(T) \leq 3.3 (2+\ln T)\sqrt{T}, ~ \forall T \geq 1.
\end{eqnarray*}
Recall that the above results hold for the pre-processed functions. Hence, scaling the bounds back by $\alpha^{-1} \equiv 2GD,$ we conclude that the proposed policy yields the following Regret and CCV bounds
\begin{framed}
\begin{eqnarray*}
 \textrm{Regret}_t &\leq& 4GD\sqrt{T}, ~~\forall t \in [T]\\ \textrm{CCV}_T &\leq& 6.6GD(2+\ln T)\big)\sqrt{T}.
 \end{eqnarray*}
 \end{framed}
\paragraph{Bounding the Movement cost}
As before, we only need to bound the quantity $\sum_{\tau=1}^T G_\tau^2$. 
We have
\begin{eqnarray*}
\sum_{\tau=1}^T G_\tau^2 &\leq& G^2\sum_{\tau=1}^T (V+ \Phi'(Q(\tau))^2 \\
&\leq& 2G^2(V^2T + n^2 \sum_{\tau=1}^T Q^{2n-2}(\tau)) \\
&\stackrel{(a)}{\leq} & 2G^2(n^{2n-2}T^n + (2n)^{2n-2}T^n) \\
&\leq & G^2 (2n)^{2n}T^n.
\end{eqnarray*}
where in (a), we have substituted the value of the parameter $V$ and used the bound from Eqn.\ \eqref{q-bd-new2}. Finally, substituting the above in \eqref{sq_movement_cost},  we conclude that the squared-Euclidean movement cost for Algorithm \ref{coco_alg} can be bounded as
\begin{eqnarray*}
		\sum_{t=1}^{T} ||x_{t+1}-x_t||_2^2 = O((\log T)^2),
\end{eqnarray*}
which is slightly larger than the logarithmic movement cost with the exponential Lyapunov function.
Using the Cauchy-Schwarz inequality, its Euclidean movement cost can also be easily bounded as \[\sum_{t=1}^{T} ||x_{t+1}-x_t||_2 \leq \sqrt{T} \sqrt{\sum_{t=1}^{T} ||x_{t+1}-x_t||^2_2}=\tilde{O}(\sqrt{T}).\]


%

\end{document}